\documentclass{article}
\usepackage[preprint]{log_2023}

\usepackage{booktabs}            
\usepackage{multirow}            
\usepackage{amsfonts}            
\usepackage{graphicx}            
\usepackage{duckuments}          
\usepackage{paralist}
\usepackage{mathtools}
\usepackage{csquotes}
\usepackage{amssymb}
\usepackage{transparent}

\usepackage[numbers,compress,sort]{natbib}

\newcommand{\pmstd}[2]{#1 {\tiny$\pm$#2}} 

\title[Filtration Surfaces for Dynamic Graph Representation]{Filtration Surfaces for Dynamic Graph Classification}

\author[F. Srambical et al.]{%
Franz Srambical \\
Helmholtz Munich \& TU Munich\\
\email{franz.srambical@helmholtz-munich.de}\And
Bastian Rieck \\
Helmholtz Munich \& TU Munich\\
\email{bastian.rieck@helmholtz-munich.de}
}

\begin{document}

\maketitle

\begin{abstract}
Existing approaches for classifying dynamic graphs either lift graph kernels to the temporal domain, or use graph
neural networks (GNNs). However, current baselines
have scalability issues, cannot handle a changing node set, or do not take edge weight information into account.
We propose \emph{filtration surfaces}, a novel method that is scalable and flexible, to alleviate said restrictions.
We experimentally validate the
efficacy of our model and show that filtration surfaces outperform previous state-of-the-art baselines on datasets that rely on edge weight
information. Our method does so while being either completely
parameter-free or having at most one parameter, and yielding the lowest overall standard deviation among similarly scalable methods.
\end{abstract}

\section{Introduction}

Recent years have seen a plethora of works dealing with the analysis of
graph-structured data \citep{veličković2018graph, xu2018how,
NEURIPS2019_9d63484a, NEURIPS2019_73fed7fd}. While the analysis of  
\emph{static} graphs is a well-studied field, the analysis of
\emph{dynamic} graphs is still nascent. In this work,
we will focus on the task of dynamic graph classification, where the
goal is to predict the class of a process, which cannot be observed from
a single snapshot of the graph, but only from the way the graph evolves
over time.
Although various methods exist for dynamic graph classification, none,
to our knowledge, concurrently offers
\begin{inparaenum}[(i)]
  \item scalability,
  \item accommodates a changing node set, and
  \item considers edge weight information
\end{inparaenum} (Table~\ref{tab:capabilitycomparison}).
In this work, we explore an approach based on 
\emph{filtrations}, a concept from topological data analysis typically 
associated with persistent homology \citep{edelsbrunner2022computational}.
Specifically, we extend prior work from 
\citet{10.1145/3447548.3467442} on \emph{filtration curves} 
to a dynamic setting.
We refer to our method as \emph{filtration surfaces}\footnote{Our code and instructions to reproduce our experiments are available at
\url{https://github.com/aidos-lab/filtration_surfaces}.}
and find that it remedies the aforementioned shortcomings of existing methods.

\section{Related Work}

The popularity of methods analysing dynamic graphs has seen a notable
surge in the last few years \citep{zhang2022dynamic, trivedi2018dyrep,
feng2022towards}.
Among these methods, a significant portion of research efforts
has been dedicated to dynamic graph neural networks~(DyGNNs), which are specifically designed to excel in link
prediction and node classification tasks~\citep{9439502}.
Although DyGNNs can handle dynamic graph classification via global
readout functions, their performance on such tasks remains largely
unexplored in the literature, with only a few isolated 
exceptions noted~\citep{MANESSI2020107000, 10.1145/3326362, 8907873}.
One such exception, GDGESN~\citep{li2023dynamical},
works by combining a Dynamical Graph Echo State Network (DynGESN)~\citep{tortorella2021dynamic} with snapshot merging, resulting in accuracies
that surpass those of the vanilla DynGESN, albeit falling short of the performance achieved by temporal graph kernels.
Another exception, STAGIN~\citep{kim2021learning}, addresses the issue of
the limited expressivity of vanilla readout functions by introducing
spatial-attention-based readout functions.
Nevertheless, a big drawback of such methods is the need for extensive
hyperparameter tuning to attain optimal performance.
\citet{wang2018time} try to solve the classification problem by first transforming a dynamic graph into a univariate or multivariate time series, and subsequently 
extracting time-series shapelets~\citep{10.1145/1557019.1557122}.
Shapelets are short time series subsequences that are maximally
discriminative between classes.
\citet{YE2023110855} employ a dynamic Dowker filtration to compute persistence
diagrams, which are vectorized to kernels and subsequently input to a support vector
machine for classification. As such, their method inherits the scalability issues of
kernel approaches.
\label{sec:temporalgraphkernels}
Finally, temporal graph kernels~\citep{doi:10.1137/1.9781611976236.56}
present an entirely different approach to dynamic graph classification
by lifting standard graph kernels to the temporal domain.
While they achieve state-of-the-art accuracies, it is important to note
that
\begin{inparaenum}[(i)]
    \item the transformation to the static graph (the~\emph{lifting}) can lead to a blow-up of the size of the graph and
    \item their computational complexity is lower-bounded by that of standard graph kernels.
\end{inparaenum}
As a consequence, the use of this method also becomes impractical for larger graphs.
By contrast, our \emph{filtration surfaces} offer a compromise that
balances computational efficiency and predictive performance.

\begin{table}
    \centering
    \caption{Comparison of the capabilities of different methods. FS-EW 
    stands for filtration surfaces with the native edge weights as the
    filter function.}
    \begin{tabular}{lcccc}
        \toprule
        Method              & Classification & Edge weights & Changing node set & Scalable \\
        \midrule
        \textit{DL-WL (SE-WL)} \citep{doi:10.1137/1.9781611976236.56}        & \textbf{yes}   & no                      & no                & no       \\
        \textit{GDGESN} \citep{li2023dynamical}             & \textbf{yes}   & no                      & no                & \textbf{yes} \\
        \textit{STAGIN} \citep{kim2021learning}             & \textbf{yes}   & \textbf{yes}            & no                & \textbf{yes} \\
        \textit{TGN} \citep{rossi2020temporal}                & no             & \textbf{yes}            & \textbf{yes}      & \textbf{yes} \\
        \midrule
        \textit{FS-EW (ours)}        & \textbf{yes}   & \textbf{yes}            & \textbf{yes}      & \textbf{yes} \\
        \bottomrule
    \end{tabular}
    \label{tab:capabilitycomparison}
\end{table}

\section{Method}
Before we introduce our proposed method, we will first provide some
background on filtrations and filtration curves. This will equip us with the necessary
tools to find a natural extension to the dynamic setting. While
filtrations are a general topological method, we restrict ourselves to
filtrations on graphs and refer the reader to \citet{edelsbrunner2022computational} for a 
more general introduction. A filtration $\mathcal{F}_G$ on a graph $G = (V, E)$ is a sequence of subgraphs $(G_1, \ldots, G_m)$ such that $\emptyset \subseteq G_1 \subseteq \ldots \subseteq G_m = G$.
A filtration can be obtained by iteratively adding edges to the starting
graph $G_1$. More specifically, a filtration can be obtained using an edge
weight function $w\colon E \rightarrow \mathbb{R}$ such that $G_i$ is induced
by all edges with weights less than or equal to $w_i$, where $w_i$ is
the $i$th smallest weight. This means that creating a filtration is
equivalent 
to sorting the edges by weight and progressively adding them to the graph in order of increasing weight, yielding a time complexity of $\mathcal{O}(n \log n)$ \citep{10.1145/3447548.3467442}.

\subsection{Filtration Curves}
\label{sec:filtrationcurves}
Filtration curves~\citep{10.1145/3447548.3467442} are expressive, 
computationally efficient representations of static graphs. Unlike methods
based on subgraph matching or neighbourhood comparison, filtration curves 
take both edge weights as well as the graph topology into account. The
fundamental postulate behind filtration curves is that two graphs generated
by a similar process have similar substructures emerging at similar
filtration timesteps.

To build a filtration curve, we need to choose
\begin{inparaenum}[(i)]
    \item an edge weight function $w\colon E \rightarrow \mathbb{R}$ that assigns a weight to every edge, and
    \item a graph descriptor function $f\colon G \rightarrow \mathbb{R}^d$ that takes a (sub)graph and returns a value in $\mathbb{R}^d$.
\end{inparaenum}
By building the graph filtration $\mathcal{F}_G$ in order of increasing edge weight and evaluating the graph descriptor function
on every subgraph $G_i$ of the filtration, one obtains a sequence of vectors which can be modeled as a matrix
$\mathcal{P}_G \coloneqq \bigoplus_{i=1}^{m} f(G_i) \in \mathbb{R}^{m \times d}$, the structure that \citet{10.1145/3447548.3467442} termed filtration curve.
In this definition, $m$ denotes the number of thresholding edge weights in the filtration $\mathcal{F}_G$, and $\bigoplus$ denotes the concatenation operator.
The filtration curve $\mathcal{P}_G$ is a compact representation of the graph $G$ that can be used for downstream tasks such as graph classification.
Another way of looking at filtration curves is as a type of (topological) feature extraction method. The graph descriptor function $f$ can be thought of as a feature extractor,
which, when evaluated alongside a filtration, yields a multi-scale representation of the graph.
Notice that the weight thresholds of $\mathcal{P}_G$ are not necessarily
equal among all graphs in the dataset.
However, it is possible to standardize the representations by creating
a shared sorted index of all weight thresholds of the dataset and
forward-filling the missing values.
\citet{10.1145/3447548.3467442} propose the following edge weight functions to define a filtration over the graph:
\begin{inparaenum}[(i)]
\item the native edge weights,
\item the max degree function $w_{xy}=\max\{\text{degree}(x), \text{degree}(y)\}$,
\item the Ollivier--Ricci curvature~\citep{OLLIVIER2009810} with $\alpha = 0.5$, and
\item the maximum of the heat kernel signatures~\citep{sun2009concise} of the adjacent nodes.
\end{inparaenum}
The Ollivier--Ricci curvature is defined as $\kappa_{\alpha}(x,y) = 1 - \frac{W(m_x^{\alpha},m_y^{\alpha})}{d(x,y)}$
for all $x,y \in V$, where $W$ denotes the Wasserstein distance~\citep{kantorovich1960mathematical} between
two probability measures, $d$ is some distance of a metric space (\citet{10.1145/3447548.3467442} use the shortest path distance), and
    $m_x^{\alpha}(v) = 
    \begin{cases}
        \alpha & \text{if } v = x \\
        \frac{1-\alpha}{\text{degree}(x)} & \text{if } v \in \mathcal{N}(x) \\
        0 & \text{otherwise,}
    \end{cases}$,
where $\mathcal{N}(x)$ denotes the set of neighbours of $x$.
The heat kernel signature is a highly expressive, but computationally
costly summary due to needing a full eigendecomposition of a matrix. \citet{carriere2020perslay} define
it as $\text{hks}(G, t, v) = \sum_{i=1}^{n} \exp(-t \lambda_i) \psi_i(v)^2$,
where $t$ is the diffusion parameter, $\lambda_i$ is the $i$th eigenvalue of the Laplacian matrix of $G$, and $\psi_i(v)$ is the $i$th eigenfunction
of the graph Laplacian. 

Besides defining a filtration over the graph, we need to choose a graph descriptor function $f$.
\citet{10.1145/3447548.3467442} propose two graph descriptor functions:
\begin{inparaenum}[(i)]
	\item counting the number of nodes with a given label, and
	\item counting the number of connected components in the graph.
\end{inparaenum}
Latter number only changes at thresholds at which a connected component
is either created or destroyed. Therefore, it suffices to only store the count at these thresholds, leading to an
even sparser representation of the graph.

\subsection{Filtration Surfaces}
Now that we have introduced filtration curves, we propose a natural way of extending them to the dynamic setting.
Specifically, we propose a method for classifying discrete-time dynamic graphs (DTDGs) $\mathcal{G}$. Intuitively, we 
calculate filtration curves $\mathcal{P}_{\mathcal{G}_i}$ for all dynamic graph timesteps $\mathcal{G}_i \in \mathcal{G}$ and therefore extend
the curve to another dimension, yielding a surface. Formally, we model the sequence of filtration curves as a tensor
$\mathcal{R}_{\mathcal{G}} \coloneqq \bigoplus_{i=1}^{n} \mathcal{P}_{\mathcal{G}_i} \in \mathbb{R}^{n \times m \times d}$,
where $n$ is the length of the dynamic graph, $m$ is the number of weight thresholds in the filtration, and 
$d$ is the dimensionality of the graph descriptor function. The careful reader will note that the filtration curves $\mathcal{P}_{\mathcal{G}_i}$
do not necessarily share the same weight
thresholds. Therefore, it is necessary to compute a shared weight index as described in section \ref{sec:filtrationcurves}.

Just like filtration curves are step functions because the graph descriptor function does not change in-between
thresholding weights, filtration surfaces can be thought of as step-like surfaces when assuming that the filtration
curve does not change in-between timestamps of the dynamic graph. Latter assumption is reasonable when the dynamic graph
is the \enquote{ground truth} and the direct result of some generation process. However, in the case of real-world
data, it is likely that the dynamic graph was obtained by sampling a streaming graph at discrete time intervals.
In such cases, forward-filling filtration curves until the next timestamp might not be the optimal approach.
Instead, our representation permits the use of any interpolation function to fill the gaps between timestamps, which
can even be learned from data.

In contrast to existing approaches, filtration surfaces can handle a changing node set, since the filtration curves are computed
independently for each graph and the shared weight index guarantees interoperability between graphs. Furthermore, our method
is suitable for the online temporal graph setting, since a new timestamp can be added to the filtration surface by simply appending
a new filtration curve to the tensor. To update the shared weight matrix, it suffices to insert the new weight thresholds
into the existing matrices and forward-fill the missing values, which
permits the use of online classifiers such as online random forests~\citep{lakshminarayanan2014mondrian}.

\paragraph{Classification}
\label{sec:classification}
To classify filtration surfaces, we vectorize them by flattening the tensor along the time dimension,
and flattening the resulting matrix along the weight dimension. The resulting vector is then input to a random forest classifier.
This means that each dimension of the input vector corresponds to a specific weight threshold at a specific timestamp. Since we
standardized all filtration curves of a dataset to have the same weight thresholds, all vectors have the same length and all vector
dimensions are comparable.

\paragraph{Limitations}
On datasets without edge weight information, filtration surfaces do not
yet obtain state-of-the-art accuracies~(Tables \ref{tab:firsttask} and
\ref{tab:secondtask}). We posit that this is largely due to the choice
of the filter and graph descriptor functions.
Since this choice is dataset-dependent, learning these functions from
data, as described in e.g.\ \citet{hofer2020graph}, constitutes an
interesting avenue for future work.

\section{Experiments \& Conclusion}
We now empirically evaluate the scalability and performance of our 
method in comparison to current state-of-the-art approaches. Our 
investigation aims to address two pivotal questions: 
\begin{inparaenum}[(i)]
    \item How does the scalability of filtration surfaces compare against
     state-of-the-art methods?
    \item How do filtration surfaces perform in relation to baselines 
    on datasets that rely on edge weight information?
\end{inparaenum}

To answer the former question, we create multiple synthetic 
datasets of varying sizes and compare the runtime as well as the representation
size of filtration surfaces against those of the current state-of-the-art: 
temporal graph kernels. We generate the datasets by constructing a
starting graph via the Barabási–Albert model~\citep{albert2002statistical}
and subsequently adding nodes and edges according to the preferential
attachment mechanism~\citep{barabasi1999emergence}. Edge weights are 
assigned randomly from a uniform distribution, either in the range
$[1, 5]$ or $[6, 10]$ depending on the class of the dynamic graph. Node
labels are assigned randomly to class $0$ or $1$. The results show that
the Gram matrices of temporal kernels scale
quadratically with the number of dynamic graphs $n$, while the corresponding filtration
surface representations scale linearly, as can be seen in Tables~\ref{tab:scalability},~
\ref{tab:scalability_part1} and \ref{tab:scalability_part2}.

To answer the latter question, we compare the classification accuracies of
filtration surfaces against those of
temporal graph kernels on our synthetic datasets. We use the
Weisfeiler--Leman kernel with the directed line graph expansion (DL-WL) and
the static expansion (SE-WL) techniques as baselines.
We use the kernel implementations from \citet{doi:10.1137/1.9781611976236.56} and
tune the hyperparameter $k$ during an inner cross-validation loop. We
evaluate the performance of all methods in a 10-fold cross-validation
setting and report the mean and standard deviation of the accuracies (Table 
\ref{tab:accuraciessynthetic}).
Filtration surfaces are able to classify the synthetic datasets perfectly,
while SE-WL and DL-WL only achieve chance accuracy.
This is due to the fact that neither method can take edge weight information
into account.

We also evaluate the performance of filtration surfaces on real-world
datasets \citep{morris2020tudataset} without edge weight information
(Tables~\ref{tab:firsttask} and \ref{tab:secondtask}). Our method (filtration
surfaces with the Ollivier-Ricci curvature as the edge weight function)
achieves the lowest overall standard deviation and competitive---albeit
not state-of-the-art---accuracies compared
to similarly scalable methods. However, in contrast to our method, none
of the comparison partners can handle a changing node set or take edge
weight information into account.

We can thus conclude that filtration surfaces constitute a robust method
with desirable scalability properties, competitive performance and the
ability to alleviate restrictions of existing approaches, making it a
novel approach in the space of dynamic graph classification methods.

\begin{table}
    \centering
    \caption{%
      Gram matrix size compared to the cumulative filtration surface size in datasets
      of varying size. OOM means that the method ran out of memory while
      having 256GB RAM allocated. Notice that the Gram matrix size
      scales quadratically with the number of dynamic graphs $n$,
      whereas the cumulative filtration surface size scales \emph{linearly}.
      }
    \begin{tabular}{lrr}
        \toprule
        $n$ & Gram Matrix Size (SE-WL) & Cumulative Filtration Surface Size (FS-EW) \\
        \midrule
        $100$ & \textbf{0.11 MiB} & 0.50 MiB \\
        $1000$ & 12.17 MiB & \textbf{4.64 MiB} \\
        $10000$ & 1312.64 MiB & \textbf{49.05 MiB} \\
        $100000$ & OOM & \textbf{491.37 MiB} \\
        \bottomrule
    \end{tabular}
  \label{tab:scalability}
\end{table}

\begin{table}
    \centering
    \caption{Accuracies~(in \%) and their standard deviations on the synthetic dataset ($n=1000$).}
    \begin{tabular}{lccc}
        \toprule
        & \multicolumn{2}{c}{WL (k=0)} & \\
        \cmidrule{2-3}
        \multirow{-2}{*}{FS-EW (ours)} & SE-WL & DL-WL & \\
        \midrule
        \textbf{\pmstd{100.00}{0.00}} & \pmstd{50.80}{4.51} & \pmstd{50.80}{4.51} \\
        \bottomrule
    \end{tabular}
    \label{tab:accuraciessynthetic}
\end{table}

\bibliographystyle{unsrtnat}
\bibliography{reference}

\appendix
\section{Appendix}

\subsection{Datasets}
We evaluate our method on five real-world datasets, namely the MIT Reality Mining dataset \citep{eagle2006reality},
the Highschool and Infectious datasets from the \emph{SocioPatterns} project \citep{sociopatterns}, the Tumblr dataset
-- a subset of the Memetracker dataset \citep{leskovec2009meme} --, as well as a subset of the Dblp dataset \citep{ley2002dblp}.
All of our datasets can be found in the \emph{TUDataset} collection \citep{morris2020tudataset}.

\citet{morris2020tudataset} obtain the datasets by generating induced subgraphs via BFS runs from each vertex. Afterwards they simulate a dissemination process
on each subgraph according to the \emph{susceptible-infected} (SI) model \citep{diekmann2000mathematical}.
The SI model is a simple epidemiological model which describes the spread of a disease in a population of susceptible and infected individuals.
At the beginning of the simulation, a random node is chosen as the starting node and labeled as infected.
In each timestep, each infected node infects each of its susceptible neighbours with probability $p$. The simulation stops
when half of the nodes are infected. The dataset statistics are shown in Table \ref{tab:datasetstatistics}.

For the first classification task, nodes of half of the dataset of dynamic graphs are labeled by simulating the SI model
with $p=0.5$, and nodes of the other half are labeled randomly. Former dynamic graphs are assigned class $0$,
while the latter are of class $1$. For the second classification task, nodes of both dynamic graph classes are labeled using the SI model
with $p=0.2$ for dynamic graphs of class $0$ and $p=0.8$ for the others. The results of the first and second classification task are shown
in Tables \ref{tab:firsttask} and \ref{tab:secondtask} respectively. 

\begin{table}[b]
    \centering
    \caption{Statistics of the datasets (from \citep{doi:10.1137/1.9781611976236.56}).}
    \begin{tabular}{clccccc}
    \toprule
    & \multicolumn{1}{c}{\multirow{2}{*}{\textbf{Properties}}} & \multicolumn{5}{c}{\textbf{Dataset}} \\ \cmidrule(lr){3-7}
    & & \textit{MIT} & \textit{Highschool} & \textit{Infectious} & \textit{Tumblr} & \textit{Dblp} \\ 
    \midrule[\heavyrulewidth]
    & Size & 97 & 180 & 200 & 373 & 755 \\
    & $\varnothing$ |V| & 20 & 52.3 & 50 & 53.1 & 52.9 \\
    & min |E| & 126 & 286 & 218 & 96 & 206 \\
    & max |E| & 3 363 & 517 & 505 & 190 & 225 \\
    & $\varnothing$ |E| & 702.8 & 262.4 & 220.4 & 98.7 & 156.8 \\
    & $\varnothing \max d(v)$ & 680.7 & 92.5 & 43.8 & 24.4 & 26.4 \\
    \bottomrule
    \end{tabular}
    \label{tab:datasetstatistics}
\end{table}

\subsection{Experimental protocol}
All experiments were conducted on a cluster, on which 8 Intel Xeon 6134 CPUs and 64 GB of RAM were allocated for each run,
except for the scalability experiments, for which 256 GB RAM were allocated.
We vectorize our filtration surfaces according to section \ref{sec:classification} and use a random forest with
1000 trees without a maximum depth as our classifier. We run 10 iterations of stratified 10-fold cross-validation
and report the mean and standard deviation of the accuracies.
For the real-world datasets, we use node label histogram filtration surfaces with
Ricci curvature as our edge weight function (FS-RC), since the dynamic graphs of the datasets do not have native edge weights.
For the synthetic case, we use node label histogram filtration surfaces with
the native edge weights as the filter function (FS-EW).

\subsection{Baselines}
Methods that start with
\emph{Stat-} are static graph classification techniques that were applied to dynamic graphs by converting them to spatiotemporal
versions. \emph{RG}, \emph{DL} and \emph{SE} \citep{doi:10.1137/1.9781611976236.56} are three approaches for converting dynamic graphs to static graphs
in a way that maximizes the preservation of temporal information. \emph{DL} and \emph{RG} lead to a bigger blow-up during the conversion than \emph{SE}.
The \emph{APPROX} methods \citep{doi:10.1137/1.9781611976236.56} are stochastic
variants of \emph{DL} with provable approximation guarantees. \emph{S} in \emph{APPROX(S=k)} denotes the number of temporal walks used for the
approximation, and RW and WL are the random walk and Weisfeiler-Leman kernels, respectively. The Graph Isomorphism Network (\emph{GIN}) and 
the Jumping Knowledge Network (\emph{JK}) are graph neural networks.

\subsection{Computational complexity}
Apart from the vectorization procedure, the computational complexity of the proposed method is that of random forests and depends on the size of the input vectors.
The input vector size is $n * m * d$. Since all trees of a random forest can be constructed in parallel, the computational complexity of
training is $\mathcal{O}(n * m * d * num\_samples * log(num\_samples))$. The dimensionality of the graph descriptor function $d$ as well as $num\_samples$ is fixed, therefore only the
number of timestamps $n$ and the number of edge weights $m$ determine the scaling behaviour. Our model thus scales linearly if the number of edge
weights is constant, and quadratically if every added dynamic graph introduces an entirely distinct set of edge weights into the dataset.
At scale, the latter behaviour is unlikely, unless the edge weights are floating point numbers. In turn, quantization of the edge weights can serve as 
an optimization to counteract unfavourable scaling behaviour in such scenarios. We leave quantization approaches for future work.

\subsection{Additional Experiments}

\begin{table}
    \centering
    \caption{Scaling metrics for SE-WL: Notice that this is the
    scaling behaviour exhibited by the most scalable
    of the methods described in
    \citep{doi:10.1137/1.9781611976236.56}, and not by DL-WL,
    another method that achieves state-of-the-art accuracies.}
    \begin{tabular}{lrrrr}
        \toprule
        $n$ & Gram Matrix & Training Time (SVM) & Inference Time (SVM) & Gram Matrix Size \\
        \midrule
        $100$ & 0.29s & 0.03s & 0.01s & 0.11 MiB \\
        $1000$ & 1.63s & 0.02s & 0.01s & 12.17 MiB \\
        $10000$ & 105.90s & 1.09s & 0.12s & 1312.64 MiB \\
        $100000$ & OOM & OOM & OOM & OOM \\
        \bottomrule
    \end{tabular}
    \label{tab:scalability_part1}
\end{table}

\begin{table}
    \centering
    \caption{Scaling metrics for filtration surfaces (FS-EW): While the cumulative
    construction times of filtration surfaces exceed those of the
    gram matrix due to our implementation storing each curve in separate files (causing
    increased I/O overhead), the scaling behavior of our method's
    construction times is clearly preferable. The I/O overhead can be
    mitigated by consolidating all curves into a single file.}
    \begin{tabular}{lrrrr}
        \toprule
        $n$ & Cum. Constr. Time & Training Time (RF) & Inference Time (RF) & Cumulative Size \\
        \midrule
        $100$ & 8.74s & 0.85s & 0.03s & 0.50 MiB \\
        $1000$ & 162.42s & 2.02s & 0.04s & 4.64 MiB \\
        $10000$ & 1391.72s & 15.95s & 0.24s & 49.05 MiB \\
        $100000$ & 5431.09s & 214.74s & 2.76s & 491.37 MiB \\
        \bottomrule
    \end{tabular}
    \label{tab:scalability_part2}
\end{table}

\begin{table}
    \centering
    \caption{Table showing classification accuracies~(in \%) and their standard deviations for the first classification task. OOM means that
    the method ran out of memory while having 64 GB RAM allocated. The
    accuracies of the comparison methods are taken from \citep{oettershagen2022temporal}.
    FS-RC denotes filtration surfaces with Ollivier-Ricci curvature as the
    edge weight function. Methods using kernels are greyed out since their scalability is not comparable to that of filtration surfaces.}
    \begin{tabular}{clccccc}
    \toprule
    & \multicolumn{1}{c}{\multirow{2}{*}{\textbf{Method}}} & \multicolumn{5}{c}{\textbf{Dataset}} \\ \cmidrule(lr){3-7}
    & & \textit{MIT} & \textit{Highschool} & \textit{Infectious} & \textit{Tumblr} & \textit{Dblp} \\ 
    \midrule[\heavyrulewidth]
    \multirow{4}{*}{\rotatebox[origin=c]{90}{Static}} & 
    \textit{\transparent{0.5} Stat-RW} & \transparent{0.5} \pmstd{61.03}{2.4} & \transparent{0.5} \pmstd{61.61}{4.3} & \transparent{0.5} \pmstd{75.80}{1.6} & \transparent{0.5} \pmstd{79.50}{1.6} & \transparent{0.5} \pmstd{83.64}{0.8} \\ 
    & \textit{\transparent{0.5} Stat-WL} & \transparent{0.5} \pmstd{43.48}{1.9} & \transparent{0.5} \pmstd{48.38}{1.5} & \transparent{0.5} \pmstd{64.95}{5.3} & \transparent{0.5} \pmstd{76.87}{0.9} & \transparent{0.5} \pmstd{78.36}{0.6} \\ 
    & \textit{Stat-GIN} & \pmstd{65.20}{4.5} & \pmstd{50.77}{5.4} & \pmstd{66.05}{3.7} & \pmstd{74.46}{2.1} & \pmstd{85.37}{1.5} \\ 
    & \textit{Stat-JK} & OOM & \pmstd{49.17}{3.7} & \pmstd{53.60}{3.8} & \pmstd{71.69}{1.7} & \pmstd{85.88}{0.7} \\ 
    \midrule
    \multirow{12}{*}{\rotatebox[origin=c]{90}{Dynamic}} & 
    \textit{\transparent{0.5} RG-RW} & \transparent{0.5} \pmstd{61.31}{2.7} & \transparent{0.5} \pmstd{90.16}{1.0} & \transparent{0.5} \pmstd{89.30}{1.0} & \transparent{0.5} \pmstd{74.99}{1.9} & \transparent{0.5} \pmstd{90.60}{1.0} \\ 
    & \textit{\transparent{0.5} RG-WL} & \transparent{0.5} \pmstd{81.88}{1.1} & \transparent{0.5} \pmstd{89.88}{0.9} & \transparent{0.5} \pmstd{91.75}{1.0} & \transparent{0.5} \pmstd{70.50}{1.0} & \transparent{0.5} \pmstd{90.45}{0.5} \\ 
    & \textit{RG-GIN} & \pmstd{50.65}{4.2} & \pmstd{51.11}{2.5} & \pmstd{58.20}{4.0} & \pmstd{72.63}{1.8} & \pmstd{86.36}{0.9} \\ 
    & \textit{RG-JK} & \pmstd{50.74}{3.1} & \pmstd{50.83}{4.9} & \pmstd{47.85}{2.7} & \pmstd{69.14}{3.6} & \pmstd{86.43}{0.7} \\ 
    & \textit{\transparent{0.5} DL-RW} & \transparent{0.5} \pmstd{92.91}{0.9} & \transparent{0.5} \pmstd{98.33}{0.7} & \transparent{0.5} \pmstd{97.05}{0.8} & \transparent{0.5} \pmstd{94.64}{0.5} & \transparent{0.5} \pmstd{98.16}{0.1} \\ 
    & \textit{\transparent{0.5} DL-WL} & \transparent{0.5} \pmstd{90.67}{1.6} & \transparent{0.5} \pmstd{98.88}{0.4} & \transparent{0.5} \pmstd{97.35}{1.5} & \transparent{0.5} \pmstd{94.05}{0.9} & \transparent{0.5} \pmstd{98.56}{0.3} \\ 
    & \textit{DL-GIN} & OOM & \pmstd{88.67}{2.1} & \pmstd{92.85}{1.7} & \pmstd{90.39}{1.4} & \pmstd{97.72}{0.4} \\ 
    & \textit{DL-JK} & OOM & \pmstd{86.22}{2.6} & \pmstd{91.55}{2.3} & \pmstd{89.30}{1.5} & \pmstd{97.57}{0.3} \\ 
    & \textit{\transparent{0.5} SE-RW} & \transparent{0.5} \pmstd{88.56}{1.0} & \transparent{0.5} \pmstd{96.89}{1.2} & \transparent{0.5} \pmstd{97.60}{0.6} & \transparent{0.5} \pmstd{93.97}{0.9} & \transparent{0.5} \pmstd{98.65}{0.3} \\ 
    & \textit{\transparent{0.5} SE-WL} & \transparent{0.5} \pmstd{87.31}{1.9} & \transparent{0.5} \pmstd{96.72}{0.7} & \transparent{0.5} \pmstd{94.45}{1.1} & \transparent{0.5} \pmstd{93.51}{0.6} & \transparent{0.5} \pmstd{97.38}{0.2} \\ 
    & \textit{SE-GIN} & \pmstd{75.98}{3.7} & \pmstd{92.28}{1.2} & \pmstd{93.10}{1.9} & \pmstd{92.78}{1.1} & \pmstd{97.87}{0.3} \\ 
    & \textit{SE-JK} & \pmstd{75.37}{3.6} & \pmstd{92.33}{2.7} & \pmstd{93.50}{1.9} & \pmstd{92.30}{0.9} & \pmstd{97.14}{0.9} \\ 
    & \textit{\transparent{0.5} APPROX (S=50)} & \transparent{0.5} OOM & \transparent{0.5} \pmstd{81.66}{1.7} & \transparent{0.5} \pmstd{84.55}{1.6} & \transparent{0.5} \pmstd{86.92}{1.2} & \transparent{0.5} \pmstd{92.56}{0.9} \\ 
    & \textit{\transparent{0.5} APPROX (S=100)} & \transparent{0.5} \pmstd{81.88}{1.0} & \transparent{0.5} \pmstd{81.66}{1.7} & \transparent{0.5} \pmstd{84.55}{1.6} & \transparent{0.5} \pmstd{86.92}{1.2} & \transparent{0.5} \pmstd{92.56}{0.9} \\ 
    & \textit{\transparent{0.5} APPROX (S=200)} & \transparent{0.5} \pmstd{83.69}{3.6} & \transparent{0.5} \pmstd{86.11}{1.2} & \transparent{0.5} \pmstd{89.35}{1.6} & \transparent{0.5} \pmstd{90.62}{0.6} & \transparent{0.5} \pmstd{94.92}{0.7} \\ 
    & \textit{\transparent{0.5} APPROX (S=500)} & \transparent{0.5} \pmstd{84.26}{3.3} & \transparent{0.5} \pmstd{91.05}{6.4} & \transparent{0.5} \pmstd{91.85}{1.7} & \transparent{0.5} \pmstd{92.73}{0.9} & \transparent{0.5} \pmstd{97.03}{0.4} \\ 
    \midrule
    & \textit{FS-RC (ours)} & \pmstd{85.08}{1.81} & \pmstd{91.88}{1.11} & \pmstd{88.70}{0.93} & \pmstd{80.49}{0.89} & \pmstd{93.65}{0.35} \\
    \bottomrule
    \end{tabular}
    \label{tab:firsttask}
\end{table}

\begin{table}
    \centering
    \caption{Table showing classification accuracies~(in \%) and their standard deviations for the second classification task. OOM means that the method ran out of memory while having 64 GB RAM allocated. Methods using kernels are greyed out since their scalability is not comparable to that of filtration surfaces.}
    \begin{tabular}{clccccc}
    \toprule
    & \multicolumn{1}{c}{\multirow{2}{*}{\textbf{Method}}} & \multicolumn{5}{c}{\textbf{Dataset}} \\ \cmidrule(lr){3-7}
    & & \textit{MIT} & \textit{Highschool} & \textit{Infectious} & \textit{Tumblr} & \textit{Dblp} \\ 
    \midrule[\heavyrulewidth]
    \multirow{4}{*}{\rotatebox[origin=c]{90}{Static}} & 
    \textit{\transparent{0.5} Stat-RW} & \transparent{0.5} \pmstd{56.84}{2.6} & \transparent{0.5} \pmstd{62.83}{2.9} & \transparent{0.5} \pmstd{63.05}{1.4} & \transparent{0.5} \pmstd{65.26}{1.9} & \transparent{0.5} \pmstd{61.17}{0.9} \\ 
    & \textit{\transparent{0.5} Stat-WL} & \transparent{0.5} \pmstd{42.42}{3.9} & \transparent{0.5} \pmstd{60.83}{3.2} & \transparent{0.5} \pmstd{63.60}{1.3} & \transparent{0.5} \pmstd{68.31}{1.5} & \transparent{0.5} \pmstd{63.11}{0.9} \\ 
    & \textit{Stat-GIN} & \pmstd{55.60}{11.0} & \pmstd{54.05}{4.7} & \pmstd{55.25}{3.3} & \pmstd{64.99}{1.1} & \pmstd{61.92}{1.2} \\ 
    & \textit{Stat-JK} & OOM & \pmstd{53.16}{3.2} & \pmstd{53.00}{2.9} & \pmstd{65.42}{2.8} & \pmstd{61.34}{1.1} \\ 
    \midrule
    \multirow{12}{*}{\rotatebox[origin=c]{90}{Dynamic}} & 
    \textit{\transparent{0.5} RG-RW} & \transparent{0.5} \pmstd{58.03}{3.7} & \transparent{0.5} \pmstd{77.33}{2.4} & \transparent{0.5} \pmstd{72.05}{2.2} & \transparent{0.5} \pmstd{68.48}{1.5} & \transparent{0.5} \pmstd{63.24}{1.2} \\ 
    & \textit{\transparent{0.5} RG-WL} & \transparent{0.5} \pmstd{66.81}{2.0} & \transparent{0.5} \pmstd{82.78}{1.3} & \transparent{0.5} \pmstd{77.40}{1.2} & \transparent{0.5} \pmstd{68.25}{1.2} & \transparent{0.5} \pmstd{66.16}{0.5} \\ 
    & \textit{RG-GIN} & \pmstd{53.80}{16.0} & \pmstd{53.61}{3.5} & \pmstd{51.80}{4.2} & \pmstd{64.70}{2.4} & \pmstd{60.24}{1.8} \\ 
    & \textit{RG-JK} & \pmstd{51.80}{9.7} & \pmstd{54.61}{2.9} & \pmstd{52.60}{3.2} & \pmstd{65.50}{2.6} & \pmstd{61.00}{1.1} \\ 
    & \textit{\transparent{0.5} DL-RW} & \transparent{0.5} \pmstd{82.64}{2.1} & \transparent{0.5} \pmstd{91.44}{1.1} & \transparent{0.5} \pmstd{87.35}{1.3} & \transparent{0.5} \pmstd{76.51}{0.5} & \transparent{0.5} \pmstd{81.79}{0.9} \\ 
    & \textit{\transparent{0.5} DL-WL} & \transparent{0.5} \pmstd{40.87}{3.6} & \transparent{0.5} \pmstd{87.11}{1.7} & \transparent{0.5} \pmstd{77.55}{2.0} & \transparent{0.5} \pmstd{78.69}{0.8} & \transparent{0.5} \pmstd{74.47}{1.1} \\ 
    & \textit{DL-GIN} & OOM & \pmstd{89.11}{2.0} & \pmstd{80.60}{2.2} & \pmstd{75.45}{2.4} & \pmstd{80.05}{1.1} \\ 
    & \textit{DL-JK} & OOM & \pmstd{85.00}{2.8} & \pmstd{75.70}{3.5} & \pmstd{73.10}{1.8} & \pmstd{79.98}{1.3} \\ 
    & \textit{\transparent{0.5} SE-RW} & \transparent{0.5} \pmstd{51.03}{5.1} & \transparent{0.5} \pmstd{90.77}{1.1} & \transparent{0.5} \pmstd{83.60}{1.1} & \transparent{0.5} \pmstd{77.09}{1.0} & \transparent{0.5} \pmstd{83.31}{1.0} \\ 
    & \textit{\transparent{0.5} SE-WL} & \transparent{0.5} \pmstd{46.52}{3.9} & \transparent{0.5} \pmstd{91.55}{0.9} & \transparent{0.5} \pmstd{79.60}{1.5} & \transparent{0.5} \pmstd{78.64}{1.4} & \transparent{0.5} \pmstd{81.24}{0.6} \\ 
    & \textit{SE-GIN} & \pmstd{51.40}{11.1} & \pmstd{85.88}{2.1} & \pmstd{75.05}{3.4} & \pmstd{73.23}{1.7} & \pmstd{80.72}{1.1} \\ 
    & \textit{SE-JK} & \pmstd{51.40}{10.9} & \pmstd{82.44}{2.0} & \pmstd{74.25}{2.4} & \pmstd{74.55}{1.4} & \pmstd{81.18}{1.0} \\ 
    & \textit{\transparent{0.5} APPROX (S=50)} & \transparent{0.5} \pmstd{55.81}{3.2} & \transparent{0.5} \pmstd{77.94}{1.8} & \transparent{0.5} \pmstd{71.70}{2.2} & \transparent{0.5} \pmstd{72.96}{1.4} & \transparent{0.5} \pmstd{68.70}{0.8} \\ 
    & \textit{\transparent{0.5} APPROX (S=100)} & \transparent{0.5} \pmstd{59.24}{5.5} & \transparent{0.5} \pmstd{83.56}{1.1} & \transparent{0.5} \pmstd{76.25}{2.5} & \transparent{0.5} \pmstd{73.03}{2.3} & \transparent{0.5} \pmstd{72.50}{0.7} \\ 
    & \textit{\transparent{0.5} APPROX (S=250)} & \transparent{0.5} \pmstd{59.48}{3.9} & \transparent{0.5} \pmstd{88.56}{1.7} & \transparent{0.5} \pmstd{78.75}{3.0} & \transparent{0.5} \pmstd{75.47}{1.2} & \transparent{0.5} \pmstd{74.44}{0.9} \\
    \midrule
    & \textit{FS-RC (ours)} & \pmstd{60.43}{1.92} & \pmstd{86.72}{0.76} & \pmstd{73.85}{1.45} & \pmstd{66.31}{0.89} & \pmstd{65.05}{0.86} \\
    \bottomrule
    \end{tabular}
    \label{tab:secondtask}
\end{table}

\end{document}